\documentclass{article}

\usepackage{microtype}
\usepackage{graphicx}
\usepackage{subcaption}
\usepackage{booktabs} 

\usepackage{hyperref}

\usepackage[preprint]{icml2026}

\usepackage{amsmath}
\usepackage{amssymb}
\usepackage{mathtools}
\usepackage{amsthm}
\usepackage{caption}
\usepackage{booktabs} 
\usepackage{tikz}
\usepackage{pgfplots}
\pgfplotsset{compat=1.18}
\usepackage{makecell}

\usepackage{graphicx}
\usepackage[dvipsnames]{xcolor}
\usepackage[capitalize,noabbrev]{cleveref}

\theoremstyle{plain}

\theoremstyle{definition}

\theoremstyle{remark}

\definecolor{PromptBG}{RGB}{248,248,248}
\definecolor{PromptFrame}{RGB}{200,200,200}
\definecolor{PromptTitle}{RGB}{40,40,40}
\usepackage[textsize=tiny]{todonotes}
\usepackage{fancyvrb} 
\usepackage[most]{tcolorbox}
\tcbuselibrary{listings, skins, breakable} 
\usepackage{fvextra} 
\fvset{breaklines=true, breakanywhere=true} 
\usepackage{listings}
\lstdefinelanguage{json}{
  basicstyle=\ttfamily\small,
  showstringspaces=false,
  breaklines=true,
  frame=single,
  rulecolor=\color{gray!50},
  upquote=true,
  literate=
   *{:}{{{\color{black}{:}}}}{1}
    {,}{{{\color{black}{,}}}}{1}
    {\{}{{{\color{blue!60!black}{\{}}}}{1}
    {\}}{{{\color{blue!60!black}{\}}}}}{1}
    {[}{{{\color{blue!60!black}{[}}}}{1}
    {]}{{{\color{blue!60!black}{]}}}}{1}
    {"}{{{\color{orange!80!black}{"}}}}{1},
  stringstyle=\color{orange!80!black},
}
\usepackage[most]{tcolorbox}
\tcbuselibrary{listings,skins,breakable}
\newtcblisting{jsonblock}[2][]{%
  listing only,
  listing options={language=json,linewidth=\linewidth,#1},
  colback=gray!3,
  colframe=gray!40,
  arc=2mm,
  left=2mm,right=2mm,top=2mm,bottom=2mm,
  boxrule=0.4pt,
  title={#2},
  breakable,
}
\newtcolorbox{promptbox}[2][]{%
  enhanced,
  breakable,
  colback=PromptBG,
  colframe=PromptFrame,
  coltitle=PromptTitle,
  fonttitle=\bfseries\small,
  title={#2},
  attach boxed title to top left={yshift=-2mm, xshift=2mm},
  boxed title style={colback=gray!10, colframe=PromptFrame, sharp corners},
  left=2mm, right=2mm, top=2mm, bottom=2mm,
  arc=2mm,
  listing only,
  listing options={style=promptstyle},
  #1
}
\usepackage{fancyvrb}
\usepackage{fvextra} 
\DefineVerbatimEnvironment{Highlighting}{Verbatim}{breaklines,commandchars=\\\{\}}

\def \our{\textsc{JudgeFlow}}
\icmltitlerunning{\our{}: Agentic Workflow Optimization via Block Judge}
\begin{document}

\twocolumn[
  \icmltitle{\our{}: Agentic Workflow Optimization via Block Judge}



\icmlsetsymbol{equal}{*}

\begin{icmlauthorlist}
  \icmlauthor{Zihan Ma}{equal,kaist}
  \icmlauthor{Zhikai Zhao}{equal,kaist}
  \icmlauthor{Chuanbo Hua}{kaist}
  \icmlauthor{Federico Berto}{kaist,radnum}
  \icmlauthor{Jinkyoo Park}{kaist,omelet}
\end{icmlauthorlist}

\icmlaffiliation{kaist}{KAIST}
\icmlaffiliation{radnum}{Radical Numerics}
\icmlaffiliation{omelet}{Omelet}
\icmlcorrespondingauthor{Zihan Ma}{zihanma@kaist.ac.kr}


  \icmlkeywords{Machine Learning, ICML}

  \vskip 0.3in
]



\printAffiliationsAndNotice{\icmlEqualContribution}

\begin{abstract}
Optimizing LLM-based agentic workflows is challenging for scaling AI capabilities. Current methods rely on coarse, end-to-end evaluation signals and lack fine-grained signals on where to refine, often resulting in inefficient or low-impact modifications. To address these limitations, we propose {\our{}}, an Evaluation-Judge-Optimization-Update pipeline. We incorporate reusable, configurable logic blocks into agentic workflows to capture fundamental forms of logic. On top of this abstraction, we design a dedicated Judge module that inspects execution traces particularly failed runs and assigns rank-based responsibility scores to problematic blocks. These fine-grained diagnostic signals are then leveraged by an LLM-based optimizer, which focuses modifications on the most problematic block in the workflow. Our approach improves sample efficiency, enhances interpretability through block-level diagnostics, and provides a scalable foundation for automating increasingly complex agentic workflows. We evaluate {\our{}} on mathematical reasoning and code generation benchmarks, where {\our{}} achieves superior performance and efficiency compared to existing methods.
\end{abstract}

\section{Introduction}

Large language models (LLMs)~\citep{brown_language_2020} have achieved remarkable success across a wide range of domains. Moving beyond the scope of foundation models~\citep{bommasani_opportunities_2022}, by integrating LLMs into intelligent agent architectures, the emerging foundation agents~\citep{liu_advances_2025} have attracted more attention. Starting from early work on prompt engineering, such as reasoning-enhanced methods~\citep{wei2023chainofthoughtpromptingelicitsreasoning,wang2023selfconsistencyimproveschainthought,yao_tree_2023}, to more recent developments in multi-agent system approaches~\citep{du_improving_2023,li_camel_2023,hong_metagpt_2024}, these handcrafted strategies have achieved strong performance across a range of tasks, including mathematical reasoning~\citep{cobbe_training_2021}, code generation~\citep{austin_program_2021}, question answering~\citep{yang2018hotpotqadatasetdiverseexplainable}, and decision making~\citep{10970024}.

However, these agentic systems still depend heavily on manual design, making workflow construction complex, costly, and inflexible. AutoML~\citep{10.5555/3360092} has shown that automating traditionally handcrafted and labor-intensive processes in machine learning can substantially reduce human effort and accelerate the development of high-performance models. Inspired by this success, recent efforts aim to automate the design and optimization of LLM-based agentic workflows~\citep{lee_compound_2025}. While these agentic systems still rely on LLMs as core execution engines, optimizing the LLMs themselves through pretraining or fine-tuning~\citep{rafailov_direct_2023} often demands substantial computational resources and massive-scale data, making such approaches expensive in many settings~\citep{kaplan_scaling_2020}. Instead, keeping the underlying model parameters fixed, and focusing on optimizing the systems structure and behavior leads to a more tractable and efficient optimization.

Automation efforts in agentic systems initially focused on prompt optimization, exemplified by Textual Gradients which leverage LLM feedback for end-to-end optimization~\citep{pryzant_automatic_2023, yuksekgonul_textgrad_2024, wang_how_2024, yin_llm-autodiff_2025}. Current efforts are expanding to optimize architecture and execution flow of entire agentic systems. Agentic workflow can be modeled as neural network~\citep{liu_dynamic_2024,ma_agentic_2025}, graph~\citep{zhuge_language_2024, zhang_multi-agent_2025}, and code~\citep{hu_automated_2025, zhang_aflow_2025, zheng_mermaidflow_2025}, each offering different levels of representational capacity, interpretability, and optimization difficulty. For instance, Directed Acyclic Graphs (DAGs)-represented workflows facilitate tractable optimization but constrain the ability to represent complex structures such as loops or conditional branching. In contrast, code-represented workflows provide comprehensive expressivity in defining intricate logic and control flow, but error attribution within code execution is difficult, and optimization often has to rely solely on end-to-end evaluation signals rather than fine-grained intermediate feedback. Building on code-represented workflows, \citet{zhang_aflow_2025} introduce operators as modular units that encapsulate common agentic actions and propose a Monte Carlo Tree Search (MCTS) framework that employs LLMs to iteratively optimize workflows using past experience. However, the expansion phase in MCTS and the subsequent evaluation of candidate workflows can be expensive, and the effectiveness of the optimization process is constrained by the granularity of guidance available for modifications. In the absence of sufficiently fine-grained diagnostic information to precisely identify which specific part within the complex workflow requires modification, the search may explore ineffective or low-impact alterations. Furthermore, complex code structural interactions such as conditional constructs where only one branch of an \texttt{if-else} statement is executed along a trajectory leave certain components without informative signals, thereby hindering fine-grained analysis.

To address these challenges, we introduce {\our{}}, an Evaluation-Judge-Optimization-Update pipeline. First, we incorporate reusable and configurable logic blocks into agentic workflows. These blocks capture three fundamental forms of logic: sequential, loop, and conditional, which are able to broadly represent code-based workflows. Compared with operators, which abstract specific agentic operations or functionalities~\citep{zhang_aflow_2025}, logic blocks serve as higher-level structural abstractions. By introducing logic blocks that abstract such common control structures, we retain the structural diversity of code-represented workflows while providing an intermediate level of abstraction between operators and workflows. This additional layer facilitates interpretability and exposes more meaningful diagnostic information for subsequent optimization.

\begin{figure}[t]
  \centering
  \includegraphics[width=0.9\linewidth]{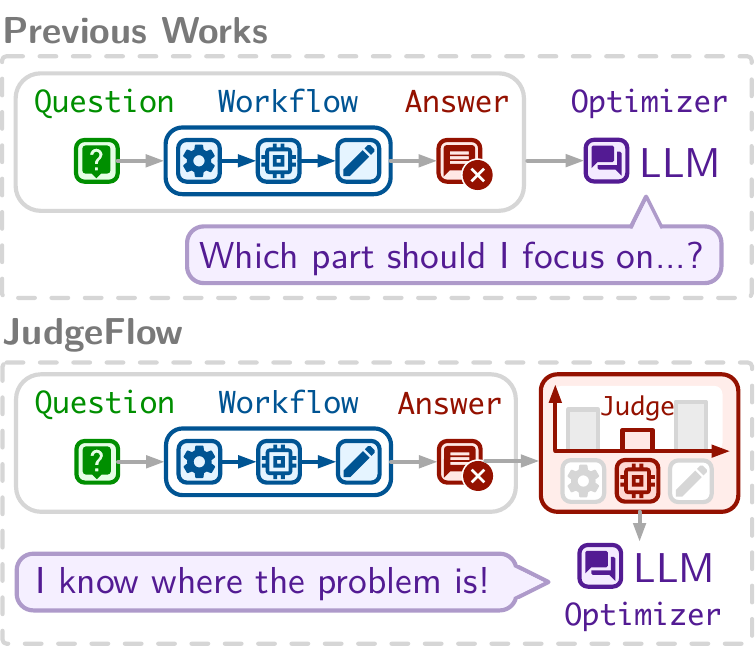}
\caption{Block-level Judge guides agentic workflow optimization by identifying the most problematic block in failed executions.}
  \label{fig:catchy}
\end{figure}

Second, we incorporate a dedicated Judge module that analyzes the execution trace, with particular emphasis on failed runs. We hypothesize that optimizers should receive both evaluation and optimization signals. For each unsuccessful execution, the Judge attempts to identify the most problematic block within the workflow as illustrated in~\cref{fig:catchy}. To further improve the precision of diagnosis, we adopt a rank-based approach at the block level. The resulting targeted diagnostic signals are propagated to the subsequent optimization stage, enabling more focused and efficient refinement of weak blocks. In this way, optimization efforts can be concentrated on repairing underperforming components, resulting in more effective and reliable improvements in overall workflow performance. Besides relying solely on end-to-end evaluation signals, our approach leverages block-level diagnostic information, enabling the optimizer to focus on the most problematic components.

In summary, our contributions are as follows: 
\begin{itemize}
\item We propose a novel Evaluation-Judge-Optimization-Update pipeline named \our{};
    \item  We introduce reusable and configurable logic blocks as higher-level structural units, which balance the expressivity of code-based workflows with tractable optimization, while supporting interpretability and intermediate execution tracing;
    \item  We design a Judge module that analyzes execution traces, especially failed runs, and assigns rank-based responsibility scores to problematic blocks, enabling fine-grained error localization and targeted refinement for subsequent optimization.
    \item We evaluate \our{} on mathematical reasoning and code generation benchmarks, showing that it outperforms existing methods. 
\end{itemize}
\begin{figure*}[t]
  \centering
  \includegraphics[width=0.95\linewidth]{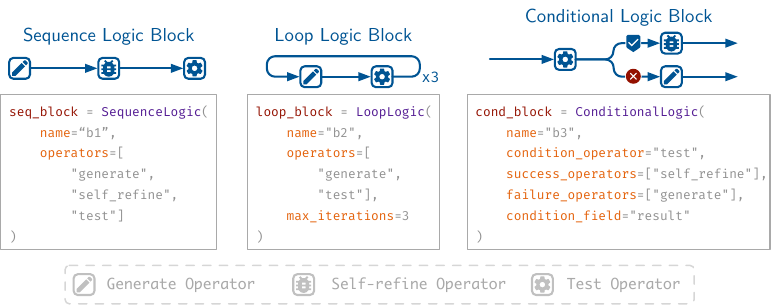}
  \caption{The illustration of logic blocks.}
  \label{fig:logic_block}
\end{figure*}
\section{Related Work}
\paragraph{LLM-based (Multi-)Agent Systems}

In recent years, LLM-based (multi-)agent systems have achieved notable successes~\citep{wang_survey_2024, huang_understanding_2024, tran_multi-agent_2025}. At the single-agent level, foundational works have enabled agents to reason and act by interleaving thought and action~\citep{yao_react_2023}, to enhance complex problem-solving through structured exploration of thoughts~\citep{yao_tree_2023}, and to interact effectively with external tools or APIs~\citep{wu_avatar_nodate}. At the multi-agent level, frameworks such as CAMEL~\citep{li_camel_2023}, AutoGen~\citep{wu_autogen_2023}, and MetaGPT~\citep{hong_metagpt_2024} have facilitated sophisticated collaboration on complex tasks, demonstrating strong performance across diverse domains. Despite these advances, existing systems remain constrained by a reliance on handcrafted prompts and rigid communication topologies, which limit adaptability as task complexity scales. This has spurred a shift toward automated agentic systems capable of optimizing their own architectures and behaviors.

\paragraph{Agentic Systems Automation} 
Early automation efforts in agentic systems primarily focused on prompt optimization~\citep{pryzant_automatic_2023, ramnath_systematic_2025, li_survey_2025}, with approaches such as LLMs-as-optimizers~\citep{yang_large_2024}, self-referential evolution~\citep{fernando_promptbreeder_2023}, textual gradients~\citep{yuksekgonul_textgrad_2024}, and self-supervised optimization~\citep{xiang_self-supervised_2025}. More recent research has expanded beyond prompt-level tuning toward optimizing the architectures and execution flows of entire agentic systems. For example, \citet{liu_dynamic_2024} explores dynamic communication structures for adaptive collaboration, while \citet{zhuge_language_2024} models agents as computational graphs to refine both prompts and inter-agent orchestration. \citet{shang_agentsquare_2024} proposes a novel modular design automatically searching for high-performance agent structures. \citet{zhou_symbolic_2024} investigates agents capable of self-optimization using symbolic optimizers. \citet{hu_automated_2025} introduces a meta agent that automatically discovers novel, high-performing, and generalizable agentic system designs. \citet{yin_goagent_2025} introduces a self-referential framework that enables agents to recursively improve themselves. \citet{zhang_aflow_2025} employs LLMs as optimizers with a Monte Carlo Tree Search (MCTS) variant to discover effective workflows. \citet{zhang_multi-agent_2025} automatically evolve agentic supernet systems leading to query-specific workflows. \citet{su_debflow_2025} leverages debate and reflexion to collaboratively refine workflows while reducing search redundancy. \citet{zheng_mermaidflow_2025} introduces safety-constrained evolutionary programming in a declarative graph space, ensuring structural validity and robustness. While these efforts mark significant progress, most existing approaches still focus on end-to-end or global architectural optimization, often leading to inefficient search and a lack of fine-grained diagnostic feedback, which limits both scalability and interpretability as task complexity grows.

\paragraph{LLM as a Judge}
The LLM-as-a-judge paradigm leverages large language models to automate the evaluation of generated content, addressing the scalability limitations of human assessment~\citep{gu_survey_2025}. This approach has been widely adopted for assessing complex outputs based on predefined criteria~\citep{li_llms-as-judges_2024}. 
To mitigate the potential bias of the LLM-as-a-Judge~\citep{wang_large_2023}, various methods have been proposed. \citet{liu_aligning_2025} propose a ranking-based alignment method that significantly improves the judging performance of LLMs. In addition, \citet{zhuge_agent-as--judge_2024} proposed the framework to use agentic systems to evaluate agentic systems. In a related application, \citet{zhang_which_2025} attempts to automate the failure attribution for LLM multi-agent systems, revealing that providing stronger ground-truth signals can substantially improve attribution quality, and aggregated analysis across multiple failures can uncover reliable error patterns.
\begin{figure*}[!t]
  \centering
  \includegraphics[width=0.9\linewidth]{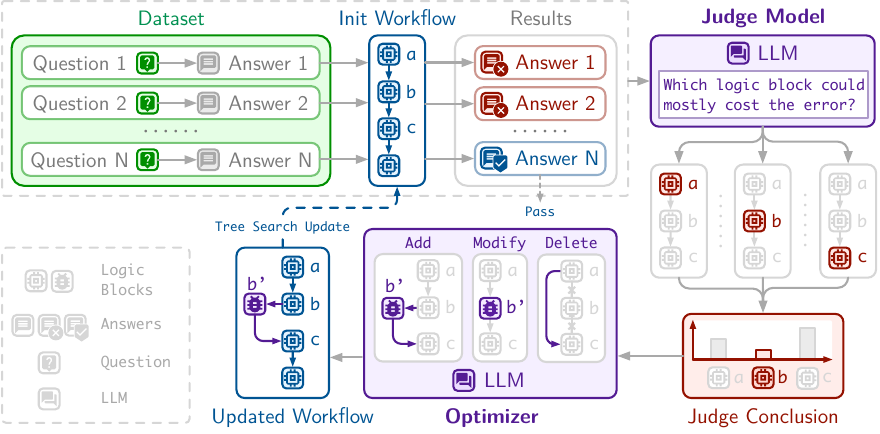}
  \caption {The main pipeline of \our{}}
  \label{fig:judgeflow}
\end{figure*}

\section{Methodology}
\subsection{Problem Formulation} 
Our framework models an agentic workflow by hierarchically composing basic agentic actions (Operators) into structured logical units (Blocks) as follows.

A configured operator $O(D)$ is the basic unit of agentic action, where $O$ represents a categorical label for its core function like \texttt{generate} or \texttt{self\_refine} (details in \cref{app:Operators}), and $D$ is the operator configuration, which includes the LLM backbone, prompt template, and other hyperparameters~\citep{zhang_aflow_2025}. Building upon operators, a logic block $(B, C)$ is a higher-level structural unit that orchestrates one or more configured operators, where $B \in \mathcal{B}$ is the logic block type, dictating how the operators are orchestrated. The set of available types $\mathcal{B}$ includes three fundamental forms of logic as shown in \cref{fig:logic_block} (details in \cref{app:LogicBlocks}):
\begin{itemize}
    \item \textbf{SequenceLogic ($\texttt{seq}$)}: A sequential execution block where operators are executed one after another. Each operator consumes the output of its predecessor, ensuring a linear flow of intermediate results until the final operator produces the block output.
    \item \textbf{LoopLogic ($\texttt{for}$)}: An iterative block that repeatedly invokes its internal operators. The iteration continues until the stopping condition is satisfied.
    \item \textbf{ConditionalLogic ($\texttt{cond}$)}: A branching block that first executes a designated condition operator. Based on the evaluation outcome, it then activates one of two operator sequences. Only the operators in the selected branch are executed to generate the block output.
\end{itemize}

Correspondingly, $C$ is the logic block configuration, which contains the set of configured operators $O(D)$ in the block and block-level parameters (e.g., stopping condition in LoopLogic). Finally, the agentic workflow $W$ is defined as a tuple $W = \left(\left\{(B_i, C_i)\right\}_{i=1}^M, S\right)$, where $M$ is the total number of logic blocks in the workflow, and $S$ denotes the ordered sequence of logic blocks at the top level while each individual block may internally contain conditional or iterative control. This definition not only preserves the common logic patterns in code-represented workflows ensuring expressive diversity~\citep{hu_automated_2025,zhang_aflow_2025}, but also enhances interpretability, including the explicit semantic characteristics of each logic block and the overall execution trajectory of the workflow facilitating subsequent optimization.

Given an input query $q$ from the dataset $\mathcal{D}$ which is available to every block, the execution function $\phi_{\text{exe}}$ processes workflow $W$ by sequentially applying its logic blocks along the execution order $S$. Each block $(B_i, C_i)$ receives the state from the previous block, $a'_{i-1}$, and produces a new state, $a'_i$, formally defined as:
\begin{equation}
a'_i = \phi_{\text{exe}}^{(i)}(a'_{i-1}, q; B_i, C_i), i=1, 2,\ldots,M,
\end{equation}
where $\phi_{\text{exe}}^{(i)}$ is the execution function for block $i$ and $a'_0=\varnothing$. The final workflow output is obtained as $a'_M$, and then scored by the evaluation function $\phi_{\text{eval}}$ against the ground-truth answer $a$ corresponding to $q$. The objective of agentic workflow optimization is to find the optimal workflow $W^*$ that maximizes evaluation performance across the dataset:
\begin{equation}
W^* = \underset{W \in \mathcal{W}}{\operatorname{argmax}}\ \mathbb{E}_{(q, a) \sim \mathcal{D}} \left[ \phi_{\text{eval}} \left( a'_M, a \right) \right],
\end{equation}
where $\mathcal{W}$ denotes the search space of candidate workflows.

\subsection{\our{}}

A challenge in optimizing agentic workflows is credit assignment: identifying which component of a complex execution is responsible for failures and should be refined. Existing methods largely rely on coarse, end-to-end evaluation signals, making it difficult to perform targeted modifications. Building on the representation of workflow using logic blocks, \our{} incorporates a dedicated Judge module and implements an iterative Evaluation-Judge-Optimization-Update pipeline for the efficient optimization of agentic workflows as shown in~\cref{fig:judgeflow}.

\subsubsection{Evaluation-Judge}
The combined Evaluation-Judge stage, detailed in \cref{alg:eval_judge}, processes each input query from the dataset. If the workflow $W$ fails on a given query, the stage identifies and logs specific problematic block within $W$. This provides targeted diagnostic signals for subsequent workflow optimization, enabling a more efficient and focused approach on refining these identified weak logic to improve overall optimization efficiency.

Specifically, for each input query $q$ (with a corresponding ground-truth answer $a$), we have $\{a_i'\}_{i=1}^M=\phi_{\text{exe}}(q, W)$, and score $s=\phi_{\text{eval}}(a_M', a)$. The score $s$ is recorded in a list $\mathcal{P}_{\text{scores}}$ for later calculation of $W$'s overall performance. Providing a threshold $\varepsilon$ that indicates successful execution, if $s \geq \varepsilon$, the instance is marked as successful, and the algorithm simply proceeds to the next input.

However, if $s < \varepsilon$, a quadruple $Q=(W, q, a, \{a_i'\}_{i=1}^M)$ is defined to encapsulate the full context of the failure. The Judge proceeds to examine the quadruple, assessing each block's $\{B_i\}_{i=1}^M$ responsibility for the failure and ranking them accordingly. This procedure, guided by specific Judge prompts (detailed in \cref{app:Judge_Prompt}), yields a rank-based score vector~\citep{liu_aligning_2025} $(r_i)_{i=1}^M$ for the blocks where $r_i=1$ refers to the block deemed most responsible for the failure and $r_i=M$ denotes the least responsible, each rank from $1$ to $M$ is assigned exactly once. These block scores $(r_i)_{i=1}^M$ are appended to $\mathcal{R}_{\text{ranks}}$. The $\texttt{RoundWorst}((r_i)_{i=1}^M, W)$ function then utilizes this score vector to identify $B_{\text{rw}}$, the block deemed most problematic for the current instance (i.e. $B_{\text{rw}} = \{\, B_i \mid r_i = 1 \,\}$) . Subsequently, the instance details $(q, a, \{a_i'\}_{i=1}^M)$ are logged into $\mathcal{L}_{B_{\text{rw}}}$, the dedicated log for $B_{\text{rw}}$, providing targeted few-shot examples for its potential future optimization.

Upon completion of all instances in $\mathcal{D}$, the accumulated diagnostic information is processed. The $\texttt{OverallWorst}(\mathcal{R}_{\text{ranks}}, W)$ function analyzes all block rank-based score vectors in $\mathcal{R}_{\text{ranks}}$ to identify $B_{\text{sel}}$, the block deemed the most consistently problematic over the whole dataset. This statistical filtering mechanism is designed to ensure robustness against noisy outputs. We follow the finding~\citep{zhang_which_2025} that while individual LLM-based failure attribution might contain noise, the aggregated distribution across multiple failures is more consistent with the true causes. In practice, we aggregate rank vectors across all failing instances in $\mathcal{R}_{\text{ranks}}$ by summing the scores $r_k$ assigned to each block $B_k$, and then selects the block achieving the minimum sum (i.e. $B_{\text{sel}} = \arg\min_{B_k \in W} \sum_{t=1}^T r_{k}^{(t)}$, where $T$ is the number of the failure executions). Concurrently, the overall performance $P_W$ of $W$ on $\mathcal{D}$ is computed by $\texttt{CalPerformance}(\mathcal{P}_{\text{scores}})$. Finally, this stage returns $P_W$, $B_{\text{sel}}$, and $\mathcal{L}_{B_{\text{sel}}}$, providing actionable insights for subsequent optimization.

\begin{algorithm}[t]
\caption{Evaluation-Judge}\label{alg:eval_judge}
\begin{algorithmic}[1]
    \STATE \textbf{Input:} Workflow $W$, Dataset $\mathcal{D}$, executor $\phi_{\text{exe}}$, evaluator $\phi_{\text{eval}}$, \text{Judge}, threshold $\varepsilon$
    \STATE \textbf{Output:} Performance $P_W$, Selected Block $B_{\text{sel}}$ and the corresponding Log $\mathcal{L}_{B_{\text{sel}}}$
    \STATE For $k \leftarrow 1 \text{ to } M$: Initialize $\mathcal{L}_{B_k} \leftarrow \emptyset$ 
    \STATE $\mathcal{R}_{\text{ranks}} \leftarrow \emptyset$, $\mathcal{P}_{\text{scores}} \leftarrow \emptyset$
    \FOR{each $(q, a) \in \mathcal{D}$}
        \STATE $\{a_i'\}_{i=1}^M \leftarrow \phi_{\text{exe}}(q, W)$
        \STATE  $s\leftarrow \phi_{\text{eval}}(a'_M, a)$ 
        \STATE $\mathcal{P}_{\text{scores}} \gets \textsc{Append}(\mathcal{P}_{\text{scores}}, s)$
        \IF{ $s\geq \varepsilon$}
            \STATE \textbf{continue} 
        \ELSE
            \STATE $(r_i)_{i=1}^M \leftarrow \text{Judge}(W, q, a, \{a_i'\}_{i=1}^M)$ 
            
            \STATE $\mathcal{R}_{\text{ranks}} \gets \textsc{Append}(\mathcal{R}_{\text{ranks}}, (r_i)_{i=1}^M)$ 
            \STATE $B_{\text{rw}} \leftarrow \texttt{RoundWorst}((r_i)_{i=1}^M, W)$ 
            \STATE $\mathcal{L}_{B_{\text{rw}}} \gets \textsc{Append}(\mathcal{L}_{B_{\text{rw}}}, (q, a, \{a_i'\}_{i=1}^M))$ 
        \ENDIF
    \ENDFOR
    \STATE $B_{\text{sel}} \leftarrow \texttt{OverallWorst}(\mathcal{R}_{\text{ranks}}, W)$ 
    \STATE $P_W \leftarrow \texttt{CalPerformance}(\mathcal{P}_{\text{scores}})$ 
    \STATE \textbf{Return:} $P_W, B_{\text{sel}}, \mathcal{L}_{B_{\text{sel}}}$
\end{algorithmic}
\end{algorithm}
\begin{table*}[h]
  \centering
  \captionsetup{width=0.8\textwidth}
  \caption{Performance comparison with baselines on \textbf{GSM8K}, \textbf{MATH}, \textbf{MBPP}, and \textbf{HumanEval}. The results are evaluated averaged over three independent runs. We use \texttt{gpt-4o-mini-0718} in the experiments.}
  \label{tab:full_width}
    \begin{tabular}{l|cccc|c}
    \toprule
    \textbf{Method} & \textbf{GSM8K}  & \textbf{MATH} & \textbf{MBPP} & \textbf{HumanEval} & \textbf{Avg.} \\
    \midrule
    \multicolumn{6}{c}{\textit{Single-agent System}} \\
    \midrule
    IO & 87.8  & 48.6 & 73.9 & 87.0 & 74.3 \\
    CoT~\citep{wei2023chainofthoughtpromptingelicitsreasoning} & 87.0  & 48.8 & 74.2  & 88.6 & 74.7 \\
    CoT SC~\citep{wang2023selfconsistencyimproveschainthought} & 86.9  & 50.4 & 73.3  & 91.6 & 75.6 \\
    \midrule
    \multicolumn{6}{c}{\textit{Hand-crafted Multi-agent System}} \\
    \midrule
    SELF-REFINE~\citep{madaan_self-refine_nodate} & 85.5  & 46.1 & 71.8  & 87.8 & 72.8 \\
    LLM-Debate~\citep{du_improving_2023} & 89.5 & 48.6 & 70.3 & 88.8 & 74.3 \\
    LLM-Blender~\citep{jiang2023llmblenderensemblinglargelanguage} & 88.4 & 46.9 & 77.1 & 88.7 & 75.3 \\
    DyLAN~\citep{liu_dynamic_2024} & 90.0 & 48.5 & 77.3 & 90.4 & 76.6 \\
    \midrule
    \multicolumn{6}{c}{\textit{Autonomous Multi-agent System}} \\
    \midrule
    GPTSwarm~\citep{zhuge_language_2024} & 89.1 & 47.9 & 77.4 & 89.3 & 75.9 \\
    ADAS~\citep{hu_automated_2025} & 88.4 & 43.2 & 77.1 & 84.2 & 73.2 \\
    AFlow~\citep{zhang_aflow_2025} & 90.1 & 52.8 & 81.7 & 90.1 & 78.7 \\
    MaAS~\citep{zhang_multi-agent_2025} & 91.5 & 52.2 & 82.2 & 91.6 & 79.4 \\
    MermaidFlow~\citep{zheng_mermaidflow_2025} & 92.4 & 55.4 & 82.3 & 92.9 & 80.8\\
    \midrule
    \textbf{\our{} (Ours)} & \textbf{93.0} & \textbf{58.5} & \textbf{83.8} & \textbf{93.4} &  \textbf{82.2}\\
    \bottomrule
\end{tabular}
\end{table*}
\subsubsection{Optimization-Update}
In the subsequent Optimization-Update stage, the LLM-based optimizer utilizes the insights from the previous stage and refines $W$ to produce an improved version $W'$ guided by specific optimization prompts (detailed in \cref{app:Optimization_Prompt}), which can be formally expressed as
\begin{equation} \label{eq:optimization_step}
W' \leftarrow \text{Optimizer}(W, B_{\text{sel}}, A, \text{sample}(\mathcal{L}_{B_{\text{sel}}}))
\end{equation}
where $\text{sample}(\mathcal{L}_{B_{\text{sel}}})$ refers to few-shot samples drawn from the logs $\mathcal{L}_{B_{\text{sel}}}$ and $A\in \mathcal{A}$, where $\mathcal{A}$ is a predefined set of available modification actions as follows:
\begin{itemize}
    \item \textbf{Add Block} : Introduce a new block $B_{\text{new}}$ with configuration $C_{\text{new}}$, and connect it directly with the low-performing block $B_{\text{sel}}$;
    \item \textbf{Remove Block}: Remove the low-performing block $B_{\text{sel}}$ together with all of its incident edges while reconnecting its predecessor and successor to preserve sequential flow;
    \item \textbf{Modify Block}: Reconfigure the existing $B_{\text{sel}}$ by updating its configuration $C_{\text{sel}} \mapsto C'_{\text{sel}}$.
\end{itemize}
In practice, the LLM-based optimizer selects $A$ adaptively based on the diagnostic signals in $\mathcal{L}_{B_{\text{sel}}}$.
Following~\citet{zhang_aflow_2025}, the refined workflow $W'$ is first evaluated to obtain its performance score $P_{W'}$. 
The pair $(W', P_{W'})$ is then added to the candidate pool $\mathcal{W}_{\text{pool}}$, which retains at most $K$ workflows by keeping the top-$K$ highest-scoring entries:
\begin{equation}
\mathcal{W}_{\text{pool}} \leftarrow \text{Top-}K\big(\mathcal{W}_{\text{pool}} \cup \{(W', P_{W'})\}\big).
\end{equation}

At the beginning of the next iteration, the optimizer selects a starting workflow $W_{\text{start}} \sim \mathcal{W}_{\text{pool}}$ according to a softmax distribution:
\begin{equation}
\Pr(W_i) = \frac{\exp\!\left(\tfrac{s_i - \max_j s_j}{\tau}\right)}{\sum_{k=1}^{|\mathcal{W}_{\text{pool}}|} \exp\!\left(\tfrac{s_k - \max_j s_j}{\tau}\right)},
\end{equation}
where $s_i$ is the evaluation score of workflow $W_i$.

\section{Experiments}
\subsection{Experimental Setups}
\paragraph{Benchmarks} We evaluate \our{} on widely used benchmarks, covering math reasoning tasks (GSM8K~\citep{cobbe_training_2021}, MATH~\citep{hendrycks_measuring_2021}, AIME~\citep{ye2025aimepreview}) and code generation tasks (MBPP~\citep{austin_program_2021}, HumanEval~\citep{chen_evaluating_2021}). Following previous studies~\citep{zhang_aflow_2025, zhang_multi-agent_2025}, each dataset is divided into training and test sets. We report the solve rate (\%) on GSM8K, MATH, AIME, and pass@1 on MBPP, HumanEval to evaluate.  

\paragraph{Baselines} We compare our \our{} with a series of baselines, including (1) Single-agent System: Standard prompting (IO), Chain-of-Thought prompting (CoT)~\citep{wei2023chainofthoughtpromptingelicitsreasoning}, and Self-Consistency~\citep{wang2023selfconsistencyimproveschainthought}; (2) Hand-crafted Multi-agent System: MultiPersona~\citep{wang2024unleashingemergentcognitivesynergy}, SELF-REFINE~\citep{madaan_self-refine_nodate}, LLM-Debate~\citep{du_improving_2023}, LLM-Blender~\citep{jiang2023llmblenderensemblinglargelanguage}, and DyLAN~\citep{liu_dynamic_2024}; (3) Autonomous Multi-agent System: GPTSwarm~\citep{zhuge_language_2024}, ADAS~\citep{hu_automated_2025}, AFlow~\citep{zhang_aflow_2025}, MaAS~\citep{zhang_multi-agent_2025}, and MermaidFlow~\citep{zheng_mermaidflow_2025}. Compared with domain-specialized agents, these methods provide a more appropriate setting to evaluate the effectiveness of our block-level Judge in workflow optimization.

\paragraph{Implementation Details} In our main experiments, to keep consistent with prior literature~\citep{zheng_mermaidflow_2025}, we use \texttt{gpt-4o-mini-0718}~\citep{openai_gpt4o_mini_2024}, \texttt{gpt-4.1-mini} \citep{openai_gpt41_mini_2025} as the optimization, Judge and execution LLM accessed via API. The number of iteration rounds is set to 20. When optimizing, we set $M \leq 3$, $\varepsilon=1$, and $K=3$. 

\subsection{Experimental Results}

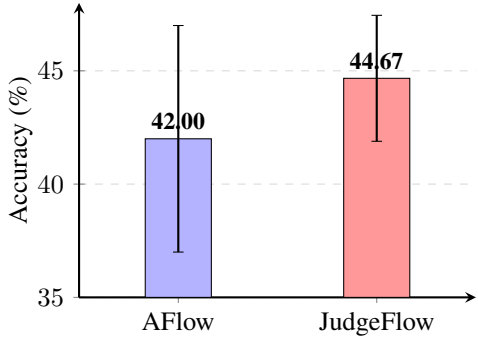
\begin{figure}
\centering
\begin{tikzpicture}
\begin{axis}[
    width=0.4\textwidth,
    height=5.5cm, 
    ymin=35, ymax=48, 
    ylabel={Accuracy (\%)}, 
    xmin=0.5, xmax=2.5,
    xtick={1,2},
    xticklabels={AFlow, JudgeFlow},
    bar width=25pt, 
    ylabel style={yshift=-5pt},
    ymajorgrids=true,
    grid style={dashed, gray!30}, 
    axis x line=bottom, 
    axis y line=left,   
    axis line style={->, >=stealth, thick}, 
]

\addplot[
    ybar,
    fill=blue!30,
    draw=black,
    error bars/.cd,
        y dir=both,
        y explicit,
        error bar style={black, thick},
]
coordinates {(1, 42.00) +- (0, 5.00)}; 

\addplot[
    ybar,
    fill=red!40, 
    draw=black,
    error bars/.cd,
        y dir=both,
        y explicit,
        error bar style={black, thick},
]
coordinates {(2, 44.67) +- (0, 2.78)}; 

\addplot[
    only marks,
    mark=none,
    nodes near coords,
    point meta=explicit symbolic,
    nodes near coords style={anchor=south, yshift=0pt, font=\small\bfseries}, 
]
coordinates {
    (1, 42.00) [42.00] 
    (2, 44.67) [44.67] 
};

\end{axis}
\end{tikzpicture}
\caption{Performance on AIME 2025. The results are evaluated averaged over five independent runs. We use \texttt{gpt-4.1-mini} in the experiments.}
    \label{fig:aime2025}
\end{figure}

As shown in~\cref{tab:full_width}, \our{} achieves superior performance compared to several strong baselines across all the tasks. Notably, for some challenging benchmarks such as MATH and MBPP, \our{} outperforms the strongest prior baseline by +3.1(5.6\%) and +1.5(1.8\%), respectively. At the same time, for relatively simpler benchmarks such as GSM8K and HumanEval, \our{} still achieves consistent gains of +0.6 and +0.5. Taken together, \our{} achieves the average score of 82.2, representing a +1.4(1.7\%) increase. As shown in \cref{fig:aime2025}, in significantly more challenging AIME benchmark, JudgeFlow achieves an average score of {44.67}. The results highlight the effectiveness of our Judge-guided block-level optimization across both reasoning and code generation tasks. 

\subsection{Analysis}
\begin{figure}[t]
\centering
\includegraphics[width=0.8\linewidth]{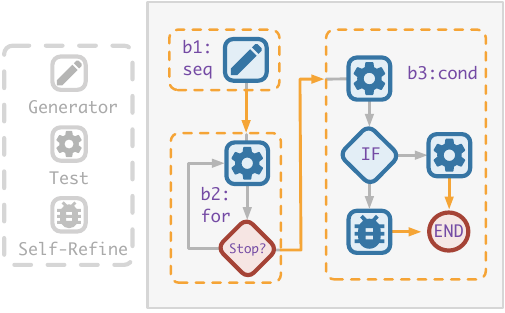}
\caption{The optimal workflow found by \textsc{JudgeFlow} on the MBPP dataset.}
\label{fig:ab1}
\end{figure}

We take the MBPP dataset as an illustrative example to analyze \our{}.
\paragraph{Best Performing Workflow} \cref{fig:ab1} is the best-performing workflow found by \our{}. First, a \texttt{seq} block \texttt{b1} applies a \texttt{generate} operator to produce an initial candidate function. Second, a \texttt{for} block \texttt{b2} repeatedly invokes the \texttt{test} operator until the stopping condition is satisfied. Finally, a \texttt{cond} block \texttt{b3} runs the \texttt{test} operator to check correctness: if the candidate doesn't pass, it routes the solution to a \texttt{self\_refine} operator.

\begin{figure}[t]
\centering
\begin{tikzpicture}
\begin{axis}[
    width=0.8\linewidth,
    height=5cm,
    xmin=1, xmax=20,
    xtick={1,5,10,15,20},
    xlabel={Iteration Number},
    ylabel={pass@1 score},
    grid=both,
    ymin=0.7, ymax=0.9,
    legend style={
        at={(0.5,1.05)},
        anchor=south,
        legend columns=2,
        nodes={scale=0.7, transform shape}
    },
]
\addplot[color=red, thick] coordinates {(1,0.7326) (2,0.7326)(3,0.7326)(4,0.7791)(5,0.8372)(6,0.8372)(7,0.8372)(8,0.8372)(9,0.8372)(10,0.8372)(11,0.8372)(12,0.8372)(13,0.8488)(14,0.8488)(15,0.8488)(16,0.8488)(17,0.8488)(18,0.8488) (19,0.8488) (20,0.8488)};
\addplot[color=red, dashed] coordinates {(1,0.739) (2,0.739)(3,0.739)(4,0.8035)(5,0.8328)(6,0.8328)(7,0.8328)(8,0.8328)(9,0.8328)(10,0.8416)(11,0.8416)(12,0.8416)(13,0.8416)(14,0.8416)(15,0.8416)(16,0.8563)(17,0.8563)(18,0.8563) (19,0.8563) (20,0.8563)};
\addplot[color=blue, thick] coordinates {(1,0.744186) (2,0.744186)(3,0.744186)(4,0.744186)(5,0.744186)(6,0.744186)(7,0.744186)(8,0.744186)(9,0.744186)(10,0.744186)(11,0.744186)(12,0.744186)(13,0.744186)(14,0.744186)(15,0.744186)(16,0.744186)(17,0.744186)(18,0.802325) (19,0.802325) (20,0.802325)};
\addplot[color=blue, dashed] coordinates {(1,0.7302) (2,0.7302)(3,0.7302)(4,0.7302)(5,0.7302)(6,0.7302)(7,0.7302)(8,0.7302)(9,0.7302)(10,0.73021)(11,0.73021)(12,0.73021)(13,0.73021)(14,0.73021)(15,0.73021)(16,0.73021)(17,0.73021)(18,0.80645) (19,0.80645) (20,0.80645)};
\legend{\textsc{JudgeFlow} (Train), \textsc{JudgeFlow} (Test), AFlow (Train), AFlow (Test)}
\end{axis}
\end{tikzpicture}
\caption{Training and testing curves of \textsc{JudgeFlow} and AFlow on the MBPP dataset.}
\label{fig:ab2}
\end{figure}
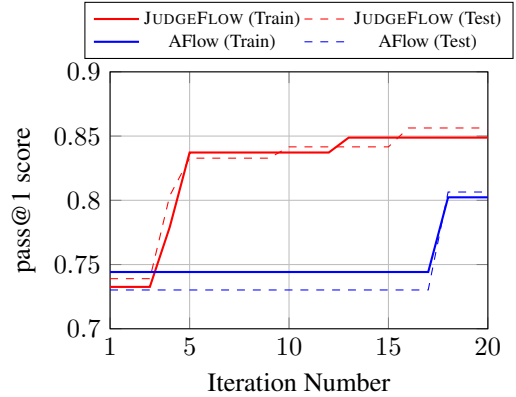

\paragraph{Ablation}
The core of \our{} is the synergy between logic blocks and the Judge module, where logic blocks provide well-defined execution boundaries that enable fine-grained failure attribution. To evaluate this design, we compare \our{} and AFlow which lacks Judge on the block-level abstraction as shown in \cref{fig:ab2}. \our{} exhibits performance gains within the first five optimization iterations, with both the training and testing curves showing rapid improvements. Beyond this early stage, \our{} continues to achieve gains, ultimately converging to higher accuracy. In contrast, AFlow remains stagnant across most iterations and only shows noticeable improvements in the later stage, and its final training and testing performance remain consistently lower than those of \our{}.

\begin{figure*}[t]
  \centering
  \includegraphics[width=0.8\linewidth]{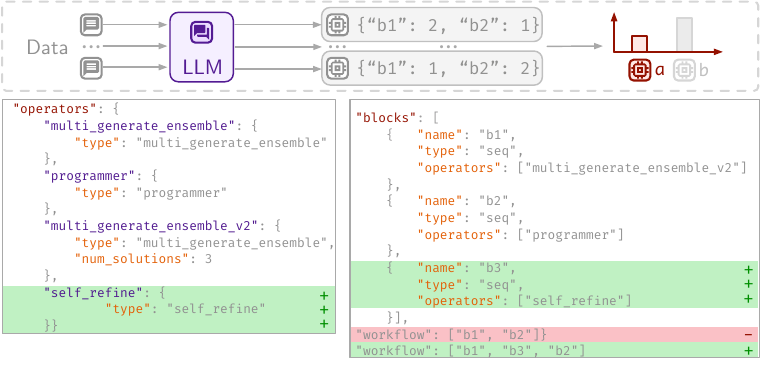}
  \caption{The illustration of the case study in the GSM8K dataset.}
  \label{fig:case-study}
\end{figure*}

\paragraph{Impact of LLMs} 
According to \cref{tab:ablation}, we keep \texttt{gpt-4o-mini-0718} fixed as the executor LLM, while varying the optimization and Judge models. Particularly, we consider \texttt{gpt-4o}~\citep{openai_gpt4o} and \texttt{Gemini-2.5-flash}~\citep{google_gemini_2.5_flash} as alternatives for these roles and report the resulting performance. The experiment confirms that increasing the capacity of optimization and Judge models consistently improves performance. While all models yield competitive results, \texttt{gpt-4o} attains the best score 84.5.

\begin{table}
\centering
\captionsetup{width=0.8\linewidth}
\caption{Performance with different LLMs on MBPP.}
\label{tab:ablation}
\begin{tabular}{lc}
\toprule
Models & Score \\
\midrule
GPT-4o-mini & 83.8 \\
GPT-4o      & 84.5 \\
Gemini-2.5-flash & 84.4 \\
\bottomrule
\end{tabular}
\end{table}

\paragraph{Cross-benchmark Generalization}

We evaluate the cross-benchmark transferability of \our{} by optimizing the workflow using the MATH(MBPP) dataset and zero-shot evaluating on the GSM8K(HumanEval) dataset. The results show that \our{} yields better transferability on both math transfer and code transfer as shown in \cref{tab:cross-generalization}.

\begin{table}[t]
\centering
\caption{Cross-Benchmark Transfer Performance}
\label{tab:cross-generalization}
\begin{tabular}{l|cc}
    \toprule
     &
    \makecell{AFlow} &
    \makecell{\our{}} \\
    \midrule
    MATH $\rightarrow$ GSM8K & 91.95 & \textbf{92.89} \\
    MBPP $\rightarrow$ HumanEval & 90.84 & \textbf{93.89} \\
    \bottomrule
    \end{tabular}
\end{table}

\paragraph{Optimization Cost} Although \our{} introduces an additional Judge module for LLM calls, the dominant cost in agentic workflow optimization lies in the Evaluation phase rather than the Judge module in our proposed methods. We monitor the cost for a single optimization round on the GSM8K dataset, the Evaluation cost (\$0.45) is considerably higher than the Judge cost (\$0.01). The Judge/Evaluation cost ratio is approximately 2\%, demonstrating that the fine-grained diagnosis provides significant optimization guidance at a marginal overhead.

\subsection{Case Study}
To illustrate how \our{} works in practice, we present a case study of workflow optimization on the GSM8K dataset as shown in \cref{fig:case-study}. The initial workflow consists of two logic blocks: \texttt{b1}, a \texttt{seq} block consisting of one \texttt{multi\_generate\_ensemble} operator designed to generate and ensemble multiple candidate solutions (with \texttt{num\_solutions} set to 3), and \texttt{b2}, a \texttt{seq} block consisting of one \texttt{programmer} operator, which takes the output from the previous block and generates the final answer using programming. When processing a batch of GSM8K instances, this workflow failed multiple times, triggering the Evaluation-Judge stage. The Judge module analyzed execution traces of these failures and assigned rank-based responsibility scores to each block. For example, in one failed run, it output \texttt{\{"b2": 1, "b1": 2\}}, attributing the primary blame to \texttt{b2}, while in another it output \texttt{\{"b1": 1, "b2": 2\}}, assigning higher responsibility to \texttt{b1}. By aggregating these rank-based scores across failures, the system identified \texttt{b1} as the \texttt{OverallWorst} block, indicating that low-quality initial solutions from \texttt{b1} were the main bottleneck, making it difficult for the workflow to generate correct final answers. In the Optimization-Update stage, the LLM-based Optimizer received this diagnostic signal and selected the \textit{Add Block} action. It introduced a new logic block, \texttt{b3}, of type \texttt{seq}, with operator \texttt{self\_refine}, which iteratively improves candidate solutions. This block was inserted between \texttt{b1} and \texttt{b2}, producing the new workflow \texttt{["b1", \textcolor{ForestGreen}{"b3"}, "b2"]}. The updated workflow first generates multiple ideas with \texttt{b1}, then refines them with \textcolor{ForestGreen}{\texttt{b3}}, and finally produces the polished answer through \texttt{b2}. This case study demonstrates how block-level diagnostics enable targeted workflow improvements.

\section{Conclusion}
In this paper, we presented a novel Evaluation-Judge-Optimization-Update pipeline named \our{} for automating agentic workflow optimization. By introducing reusable logic blocks as higher-level structural abstractions, \our{} achieves a balance between the expressive flexibility of code-based workflows and the tractability of optimization. On top of this representation, the Judge module provides block-level diagnostic signals by analyzing execution traces and assigning responsibility to the problematic block, enabling more interpretable and fine-grained optimization. Through extensive experiments on mathematical reasoning and code generation benchmarks, we demonstrate that \our{} consistently outperforms strong baselines.
Future work may include exploring more robust Judge for agentic systems optimization.








\section*{Impact Statement}

This paper presents work whose goal is to advance the field of Machine
Learning. There are many potential societal consequences of our work, none
which we feel must be specifically highlighted here.

\bibliography{example_paper}
\bibliographystyle{icml2026}

\newpage
\appendix
\onecolumn
\section{Operators}
\label{app:Operators}
Following~\citet{zhang_aflow_2025},~\citet{zhang_multi-agent_2025} and~\citet{zheng_mermaidflow_2025}, we adopt the following set of operators:
\begin{enumerate}
    \item \texttt{generate}, 
    a generation operator that produces candidate solutions based on the problem description and optional previous results.

    \item \texttt{test}, 
    a testing operator that executes generated solutions against test cases and provides feedback for refinement.

    \item \texttt{self\_refine}, 
    a refinement operator that improves a given solution through self-refinement.

    \item \texttt{multi\_generate\_ensemble}, 
    an ensemble operator that generates multiple solutions and combine them to the best one via self-consistency.

    \item \texttt{programmer}, 
    a synthesis-and-execution operator that generates Python code for solving math problems, runs it in a restricted environment, and iteratively repairs errors.
\end{enumerate}

\section{Logic Blocks}
\label{app:LogicBlocks}
We implement three common logic types in code-represented workflows: \textbf{SequenceLogic (\texttt{seq})}, \textbf{LoopLogic (\texttt{for})}, and \textbf{ConditionalLogic (\texttt{cond})}, whose descriptions and interfaces are listed below.
\begin{jsonblock}{Logic Blocks}
{
    "SequenceLogic": {
        "type": "seq",
        "description": "Execute operators strictly in order. Required fields: name (string), type (must be 'seq'), operators (array of operator aliases). No optional fields. Use this for linear processing flows where you need sequential execution of operators.",
        "structure": {
            "name": "block_name",
            "type": "seq",
            "operators": ["operator"]
        },
        "input_flow": "block_input -> op1 -> op2 -> ... -> block_output"
    },
    "LoopLogic": {
        "type": "for",
        "description": "Iteratively execute a sequence of operators until the optional asynchronous condition returns False or the max iteration limit is reached. Required fields: name (string), type (must be 'for'), operators (array of operator aliases). Optional fields: max_iterations (integer, default 3), condition (object with 'field' and 'equals' properties, or null for no condition). Use this for retry mechanisms and iterative refinement.",
        "structure": {
            "name": "block_name",
            "type": "for",
            "operators": ["operator"],
            "max_iterations": num_iterations,
            "condition": {
                "field": "field_name",
                "equals": "some_value" 
            }
        },
        "input_flow": "block_input -> repeat [op1 -> op2 -> ...] until stop -> block_output"
    },
    "ConditionalLogic": {
        "type": "cond",
        "description": "Run a dedicated condition operator first, then choose the success or failure branch based on the field specified by 'condition_field'. The chosen branch runs sequentially with the same data-passing semantics as SequenceLogic. Required fields: name (string), type (must be 'cond'), condition_operator (string, operator alias to evaluate condition), success_operators (array of operator aliases for success path), failure_operators (array of operator aliases for failure path). Optional fields: condition_field (string, field name to check for condition result, default 'result'). The condition operator evaluates criteria and sets a result field, which determines whether to execute success_operators or failure_operators. Use this for branching logic and conditional processing. ",
        "structure": {
            "name": "block_name",
            "type": "cond",
            "condition_operator": "condition_operator",
            "success_operators": ["success_op"],
            "failure_operators": ["failure_op"],
            "condition_field": "field_name"
        },
        "input_flow": "block_input -> condition operator -> select branch -> branch sequence -> block_output"
    }
}

\end{jsonblock}

\section{Judge Prompt}
\label{app:Judge_Prompt}

\begin{promptbox}{System Prompt}
\begin{Verbatim}[fontsize=\small]
You are a workflow failure analyst. Given execution evidence from a block-based AI workflow that produced an incorrect answer, determine which logic block is causally responsible for the failure.

# Knowledge Base
## Logic block types
{logic_block_descriptions_text}

## Operator types
{operator_descriptions_text}

# Responsibility Principles:
- Consider blocks that actually make mistakes over blocks that only perform redundant work.
- Our goal is to identify the weakest block in this workflow, so that in later optimization we can focus on improving this weakest block.
- You will be given: the problem, the correct answer, the incorrect answer, the workflow execution trace, and each block's inputs/outputs in a sequential pipeline. Ground your judgment in this evidence:
    - For each block, compare its output vs. input, and output vs. the correct answer to locate where the first critical deviation was introduced, how later blocks propagated/amplified it, and whether any block had enough information to correct it but failed to do so.
    - Do not overweight temporal order:
    - Earlier blocks bear more responsibility for introducing the critical error.
    - Later blocks bear responsibility for failing to correct earlier errors given the available context.
- If two blocks seem equally responsible, apply counterfactual reasoning: If this block were correct, would the final answer be correct? 
- You may form a brief internal natural-language reason (e.g.,"this block generated incorrect code") to aid the decision, but the output must be JSON only.

# Output Contract
Return a JSON object mapping each block name to a unique integer rank (1 = most responsible, n = least responsible). Each rank from 1 to n must appear exactly once. Output JSON only, no explanations.

\end{Verbatim}
\end{promptbox}

The user prompt provides the problem, correct answer, incorrect answer, workflow structure, execution trace in XML format, and the list of blocks to rank.

\section{Optimization Prompt}
\label{app:Optimization_Prompt}

\begin{promptbox}{System Prompt}
\begin{Verbatim}[fontsize=\small]
You are an expert workflow optimization assistant specializing in Logic Block-based AI workflows for the {{dataset}} dataset.

IMPORTANT: Focus exclusively on optimizing the low-performing logic block to improve code generation quality and overall workflow performance.
IMPORTANT: You have exactly one optimization attempt. Reason carefully and aim to improve performance across the entire dataset.

# Task Overview

You will be provided with:
1. Error examples showing: problem, correct answer, workflow's wrong answer, and the low-performing block's output
2. Current workflow definition 
3. Performance analysis results

Your objective: Optimize the identified low-performing logic block using the error examples as guidance while avoiding overfitting.

# Logic Block Types and Detailed Semantics
{logic_blocks_section}

# Available Operators
{operators_section}

# Critical Instructions for Operator Usage

INSTRUCTION Field is Crucial: 
- The `instruction` field is extremely important for operator performance and directly impacts final output quality
- Instructions should clearly guide the operator on how to process input and produce expected output
- For code generation tasks, instructions need to include specific programming requirements, output format, and quality standards
- For mathematical reasoning tasks, instructions need to include specific problem-solving approaches, step-by-step reasoning requirements, and output format standards

# Optimization Strategies

Choose exactly one strategy:

## 1. Add Block Strategy
- Create a completely new logic block with its own name (e.g., "b2", "b3")
- Insert the new block immediately before or after the low-performing block
- Select appropriate block type (seq/for/cond) that complements the low-performing block
- Populate all required parameters (instructions, iteration limits, condition fields, etc.)
- Run internal counterfactual reasoning but do not output explorations

Example: from `"workflow": ["b1", "b2"] ("b2" performs worst) to "workflow": ["b1", "b2", "b3"]`

## 2. Remove Block Strategy  
- Completely delete the low-performing block when it adds noise or harms outcomes
- Internally evaluate workflow behavior without that block
- Update workflow sequence and remove unused operators

Example: from `"workflow": ["b1", "b2"] ("b1" performs worst) to "workflow": ["b2"]`

## 3. Modify Block Strategy
- Rework the existing low-performing block without introducing new blocks
- Examine block's logic type, operator choices, and parameterization
- Update operators, ordering, and configuration for stronger reasoning
- Focus solely on refining the current block

# Critical Constraints

CRITICAL: Maximum 3 blocks per workflow - DO NOT EXCEED this limit
CRITICAL: Create NEW BLOCK with different name when adding
IMPORTANT: Focus on the low-performing block identified in the analysis
IMPORTANT: Maintain compatibility with other blocks in the workflow
IMPORTANT: Each block should have a clear, distinct purpose

# Prohibited Actions

- NEVER reproduce workflow configurations matching provided history
- MUST NOT repeat, reuse, or recycle any optimization from Previous Optimization Analysis
- All workflows in previous optimization analysis are explicitly banned
- Run internal "novelty check" to confirm at least two structural differences from banned workflows

# Output Requirements

- Apply exactly one modification strategy (Add/Remove/Modify)
- Focus only on the identified low-performing logic block  
- Output clean JSON without comments or explanations
- Ensure JSON is fully parseable and syntactically correct
- Avoid overfitting to provided error examples
\end{Verbatim}
\end{promptbox}

\begin{promptbox}{User Prompt}
\begin{Verbatim}[fontsize=\small]

## Dataset
<dataset>{dataset}</dataset>

## Current Workflow Performance
Current workflow score: <score>{score}</score>

Low-performing logic block identified:
<low_performing_blocks>{low_performing_blocks}</low_performing_blocks>

## Current Workflow Definition
```json
<previous_code>{previous_code}</previous_code>
```

## Error Analysis
Error examples show:
- Problem: Original code generation task/question
- Correct Answer: Expected output
- Workflow Wrong Answer: Current workflow output
- Low-performing Block Output: Problematic block's specific output

## Previous Optimization History
STRICTLY PROHIBITED: Do not repeat or reuse any optimization results below.
<reflection_result>{reflection_result}</reflection_result>

IMPORTANT: All workflows above and current definition are disallowed baselines.

# Optimization Task

Analyze the low-performing logic block and improve its output quality.

## Core Optimization Objective
Your optimization purpose is to modify the weakest block:
- Deeply analyze why this weak block led to the final incorrect answer
- Understand the block's role and impact within the entire workflow
- Identify the specific failure patterns and root causes of this block
- Your chosen action (Add/Modify/Remove) should be aimed at solving the current problems

## Key Focus Areas
- Low-performing block is your primary optimization target
- Use error cases to understand failure patterns
- Improve block's reasoning or processing capability
- Evaluate block type appropriateness (seq/for/cond)
- Assess operator suitability and configuration
- Pay special attention to the quality and detail of instruction fields

## Strategy Guidelines
Current workflow has <workflow_block_count>{workflow_block_count}</workflow_block_count> block(s).

## Error Examples
Use these to understand failures, but avoid overfitting:
<error_cases_section>{error_cases_section}</error_cases_section>

\end{Verbatim}
\end{promptbox}


\end{document}